\documentclass[runningheads,a4paper]{llncs}

\usepackage{amssymb}
\usepackage{graphicx}
\usepackage{amsmath}
\usepackage{url}
\usepackage{tabularx}

\usepackage[colorlinks,
            linkcolor=red,
            anchorcolor=blue,
            citecolor=green
            ]{hyperref}

\usepackage[style=nature,hyperref=true,backend=biber,sorting=none]{biblatex} 
\urldef{\mailsa}\path|yangsonglin2021@ia.ac.cn, wwang@nlpr.ia.ac.cn *,|
\urldef{\mailsb}\path|chengyuehua@nuaa.edu.cn, jdong@nlpr.ia.ac.cn|
\newcommand{\keywords}[1]{\par\addvspace\baselineskip
\noindent\keywordname\enspace\ignorespaces#1}

\begin{document}

\mainmatter  % start of an individual contribution

\title{A Systematical Solution for Face De-identification}

\titlerunning{}

\author{Songlin Yang\textsuperscript{1,}\textsuperscript{2}, Wei Wang\textsuperscript{1*}, Yuehua Cheng\textsuperscript{2}
\and Jing Dong\textsuperscript{1}}
\authorrunning{ }

\institute{\textsuperscript{1}Chinese Academy of Sciences, Institute of Automation, \\
Center for Research on Intelligent Perception and Computing, Beijing, China\\
\textsuperscript{2}Nanjing University of Aeronautics and Astronautics, \\College of Automation, Nanjing, China\\
\mailsa\\
\mailsb
}


\toctitle{Lecture Notes in Computer Science}
\tocauthor{Authors' Instructions}
\maketitle

\begin{abstract}
With the identity information in face data more closely related to personal credit and property security, people pay increasing attention to the protection of face data privacy. In different tasks, people have various requirements for face de-identification (De-ID), so we propose a systematical solution compatible for these De-ID operations. Firstly, an attribute disentanglement and generative network is constructed to encode two parts of the face, which are the identity (facial features like mouth, nose and eyes) and expression (including expression, pose and illumination). Through face swapping, we can remove the original ID completely. Secondly, we add an adversarial vector mapping network to perturb the latent code of the face image, different from previous traditional adversarial methods. Through this, we can construct unrestricted adversarial image to decrease ID similarity recognized by model. Our method can flexibly de-identify the face data in various ways and the processed images have high image quality.
\end{abstract}

\section{Introduction}

The privacy of biometrics becomes increasingly important, with face recognition widely used. Face data contains important identity information, which is closely related to personal credit and property security. Face data is collected more and more frequently, and people tend to show photos on the Internet, which can be easily obtained by others. As lots of abuse of face data is exposed, people pay increasing attention to the security of face data privacy.

The traditional face anonymization, such as mosaic or blurring of face region, has the disadvantage of degrading the quality of the images significantly. These operations destroy the original data distribution and reduce the detection rate in exchange for the reduction of recognition rate, which makes users or data mining researchers unable to further effectively use anonymous face data.

With the concepts of adversarial attacks\cite{szegedy2013intriguing} and generative adversarial networks\cite{goodfellow2014generative} proposed, many related works have been published, such as Fawkes\cite{shan2020fawkes} and CIA-GAN\cite{maximov2020ciagan}. However, there are still some unsolved problems in these methods: the perturbations of random noise will lead to the degradation of image quality, and the face swapping will generate some artifacts which are easily perceived by human eyes.

To meet the various needs reflected in the above description, our method of face de-identification is proposed. On the one hand, it can realize the de-identification for human visual perception, i.e., completely changing the identity by face swapping. This can meet the requirements of high-intensity anonymization, for De-ID preprocess of face dataset or removing the identification completely when the information is shown to the public. On the other hand, our method can realize the de-identification for face recognition model, i.e., using the adversarial examples to perturb the identification results of the model. Therefore, for an individual user, face privacy data protection is carried out without the loss of original information, while his sharing needs are satisfied. For example, the De-ID images can be displayed on social platforms, which can be recognized by his friends without being collected identity information maliciously.

\textbf{The contributions of this paper can be summarized as follows:}

\textbf{a.} A systematical solution for face de-identification is constructed, which can De-ID for the human visual perception by face swapping,  and De-ID for face recognition model by adversarial attacks.

\textbf{b.} Our method perturbs the latent vector of the image in the latent space and maps this vector to the latent space of StyleGAN\cite{karras2019style}, to get adversarial images with high quality.

\textbf{c.} Different from traditional adversarial methods which add noise to the original image, our method propagates the gradient back to the mapping network of the adversarial vector. This is to explore a new universal and unrestricted adversarial sample generation\cite{song2018constructing} way for face images.

\section{Background and related work}
\subsubsection{De-identification (De-ID)}
Considering different needs of privacy protection, de-identification has the following two definitions: one is for human visual perception, with the purpose of completely hiding the identity information of the original image. This is mainly for the applications of public display (complete anonymization) and dataset desensitization (massive data anonymization), represented by mosaic, blurring and face swapping. The other is for face recognition model. In order to prevent the abuse of identity information, it is mainly aimed at the privacy protection of individual users, under the premise of meeting the sharing needs, represented by adversarial examples.
\subsubsection{Attribute disentanglement} 
Many works such as FaceShifter\cite{li2019faceshifter} and StyleGAN, connect the visual facial attributes with high-dimensional vectors, which can enhance the ability of controlling the generative network. For the De-ID problem, we put forward higher requirements for attribute disentanglement. We hope that the information used by the models to recognize human identity can be decoupled from other facial features, such as eyes, nose, jawline, etc., so that the De-ID problem can be solved. These works\cite{chen2016infogan}\cite{higgins2016beta} have made some contributions, but they still cannot meet the needs of De-ID tasks.
\subsubsection{StyleGAN}
Inspired by the paper\cite{nitzan2020face}, our method takes the pre-trained synthesis network of StyleGAN as the generator. We train a mapping network to map the latent vector of other latent spaces to the latent space of StyleGAN, which inherits its capacity of generating highly realistic and high quality images. Through this operation, we can split the attribute disentanglement and image generation, to avoid the degradation of image quality.
\subsubsection{Adversarial attacks on face recognition systems}
Bose et.al.\cite{bose2018adversarial} make the faces cannot be detected by detectors, while others use patches\cite{kaziakhmedov2019real}, glasses\cite{sharif2016accessorize}, hats\cite{komkov2019advhat} and other methods to realize physical adversarial attacks. However, these methods are not suitable for the De-ID preprocess of large datasets. Our method will explore the universal adversarial features of the data, and get a network that can describe the common patterns of the adversarial samples corresponding to each class of images.

\begin{figure}
\setlength{\abovecaptionskip}{0cm} 
\setlength{\belowcaptionskip}{0cm}
\centering
\includegraphics[scale=0.34]{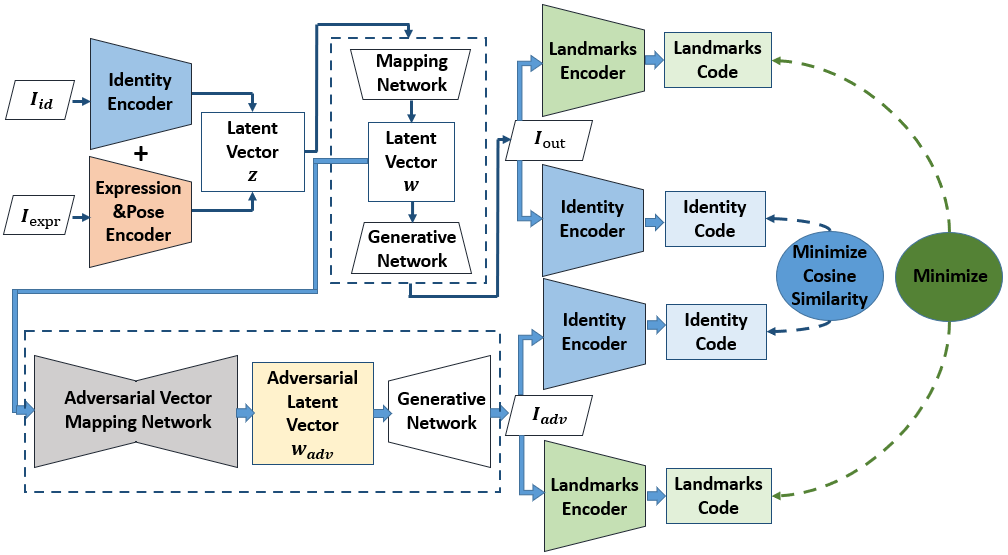}
\caption{Framework of our algorithm.}
\label{Figure 1}
\end{figure}

\section{Method}

\subsection{Attribute disentanglement and generative network}

As shown in Fig.\ref{Figure 1}, the identity encoder ($En_{id}$) extracts the code of facial features from $I_{id}$. The expression and pose encoder ($En_{expr}$) extracts the code of expression and pose from $I_{expr}$. The latent vector $z$ is combined by these two codes, and feed it into the mapping network $M$ to obtain the latent vector $w$, which will be sent into the generative network to output the $I_{out}$, thus completing the fusion of identity and expression from different faces. Therefore, the function of \textbf{different-input-and-face-swap} and \textbf{same-input-and-face-reconstruct} can be realized.

\subsubsection{Network architecture}

The ResNet-50 face recognition model, which is pre-trained on VGGFace2 dataset, is used as $En_{id}$ to extract the information of identity,  while using the Inception-V3 network as the $En_{expr}$ to extract information of expression and pose. The outputs of both encoders are taken from their last feature vector before the fully-connected classifier and the latent vector $z$ is obtained by combining the outputs of the two encoders:

\begin{equation}
    z=[En_{id}(I_{id}),En_{expr}(I_{expr})]
\end{equation}

The mapping network is a four-layers MLP using LReLU as activation layers, which maps the latent vector $z$ to the latent vector $w$, thus satisfying the distribution of the generative network. The generator is the synthesis network of StyleGAN. Feed the latent vector $w$ into the generator and get the output image. Landmarks encoder ($En_{lnd}$) is implemented using a pre-trained landmarks regression network and we only use the 52 inner-face landmarks, removing the jawline landmarks. Discriminator $D_{W}$ is used to judge whether the latent vector $w$ obtained by mapping network satisfies the distribution of generative network.

\subsubsection{Loss function} 

The loss functions can be divided into adversarial loss and non-adversarial loss. For adversarial loss, the non-saturating loss with $R_{1}$ regularization is selected:

\begin{equation}
    \begin{split}
        L_{adv}^{D}=
        &-E_{w\sim W}[log(D_{W}(w)]-E_{z}[log(1-D_{W}(M(z)))]\\
        &+\frac{\gamma}{2}E_{w\sim W}\|\nabla_{w}D_{W}(w)\|_{2}^{2}
    \end{split}
    % L_{adv}^{D}=-E_{w\sim W}[log(D_{W}(w)]-E_{z}[log(1-D_{W}(M(z)))]+\frac{\gamma}{2}E_{w\sim W}\|\nabla_{w}D_{W}(w)\|_{2}^{2}
\end{equation}

\begin{equation}
    L_{adv}^{G}=-E_{z}[log(D_{W}(M(z)))]
\end{equation}

For the non-adversarial loss, the overall generator non-adversarial loss $L_{non-adv}^{G}$ is a weighted sum of $L_{id}$, $L_{landmark}$ and $L_{reconstruct}$:

\begin{equation}
    L_{non-adv}^{G}=\lambda_{1}L_{id}+\lambda_{2}L_{landmark}+\lambda_{3}L_{reconstruct}
\end{equation}

As known to all, human perception is highly sensitive to minor artifacts, because not only does every individual frame must look realistic, but the motion across frames must also be realistic. 

To enforce visual identity preservation, we use $L_{1}$ cycle consistency loss between $I_{id}$ and $I_{out}$:

\begin{equation}
    L_{id}=\|En_{id}(I_{id})-En_{id}(I_{out})\|_{1}
\end{equation}

To model the possible motion of human face better, we use $L_{2}$ cycle consistency landmarks loss between $I_{expr}$ and $I_{out}$:

\begin{equation}
    L_{landmark}=\|En_{lnd}(I_{expr})-En_{lnd}(I_{out})\|_{2}
\end{equation}

To encourage pixel-level reconstruction of the image when $I_{id}$ and $I_{expr}$ are the same, we use the ‘mix’ loss suggested by Zhao et.al.\cite{zhao2016loss}, and use a weighted sum of $L_{1}$ loss and MS-SSIM loss. Furthermore, to prevent this reconstruction loss to effect the training of disentanglement, we only employ the reconstruction loss when $I_{id}$ and $I_{expr}$ are the same:

\begin{equation}
    L_{reconstruct}=
    \left\{
         \begin{array}{lr}
         L_{mix}, & I_{id}=I_{expr} \\
         0, & I_{id}\not=I_{expr}
         \end{array}
    \right.
\end{equation}

\begin{equation}
    L_{mix}=\alpha(1-(MS-SSIM(I_{id},I_{out})))+(1-\alpha)\|I_{id}-I_{out}\|_{1}
\end{equation}

We sample 70K random Gaussian vectors and feed them through pre-trained StyleGAN, and then we get the latent vector $w$ and its corresponding generative image. The $(image,w)$ data is randomly sampled from the training dataset, and the sampled image data is used as the input of $I_{id}$ and $I_{expr}$, while the latent vector $w$ will be used as the positive sample for training discriminator. The cross-face training and the same-face training are carried out alternately according to the frequency. The parameters of $En_{expr}$, discriminator $D_{W}$ and mapping network $M$ will be updated after each round of training. Note that we separately update the adversarial and non-adversarial parts of parameters, to make the process of training more stable.

\subsection{Adversarial vector mapping network}

As shown in Fig.\ref{Figure 1}, take the latent vector $w$ in the pipeline of attribute disentanglement and generative network, and feed it into the adversarial vector mapping network. Then get the adversarial latent vector $w_{adv}$, and feed it into the generative network. Finally, we can obtain the adversarial image $I_{adv}$ which ID similarity recognized is low against the target ID, but looks very similar to the original image, thus perturbing the identical results of face recognition systems.

\subsubsection{Network architecture}

Based on the attribute disentanglement and generative network, our method adds an adversarial vector mapping network $M_{adv}$ into the framework. This is a four-layers MLP using LReLU as activation layers, which maps the latent vector $w$ to the adversarial latent vector $w_{adv}$.
    
\subsubsection{Loss function}

To better maximize the inner-class distance between the original image and the adversarial image in the latent space, we use the cosine similarity to measure the distance between both images:

\begin{equation}
    L_{id}^{model}=\frac{En_{id}(I_{id})\cdot En_{id}(I_{adv})}{\|En_{id}(I_{id})\|\cdot\|En_{id}(I_{adv})\|}
\end{equation}

Note that we have to restrict the cosine similarity between the original image and the adversarial image in the latent space, and clip $w_{adv}$ by some selected value $\delta$, after the adversarial vector mapping network outputs the $w_{adv}$:

\begin{equation}
    w_{adv}=Clip_{\delta}(w_{adv},w)
\end{equation}

At the same time, we expect that the human perception similarity between the original image and the adversarial image is high, so we minimize the $L_{2}$ cycle consistency distance between landmark codes of these two images:

\begin{equation}
    L_{id}^{visual}=\|En_{lnd}(I_{expr})-En_{lnd}(I_{adv})\|_{2} 
\end{equation}

Finally, we adopt a weighted sum of two loss functions as the optimization objective:

\begin{equation}
    L_{adv}=\lambda_{4} L_{id}^{model}+\lambda_{5} L_{id}^{visual} 
\end{equation}

Note that we only use the same-face training in this training stage. Calculate the gradients of the adversarial vector mapping network and update the parameters of it after getting the adversarial latent vector $w_{adv}$.

\section{Experiment}

This section will evaluate the performance of the two parts of our method qualitatively and quantitatively, from the ability of disentanglement, the quality of generative images, the adversarial performance of adversarial images and so on. We use FGSM\cite{goodfellow2014explaining}, ALAE\cite{pidhorskyi2020adversarial} (generates the whole image) and FSGAN\cite{nirkin2019fsgan} (generates the region of face) as the comparisons. We use the FID\cite{heusel2017gans} (Fréchet Inception Distance) as the objective metric to measure the similarity between the original image and the generative image. We choose MTCNN\cite{zhang2016joint} and Arcface\cite{deng2019arcface} as face recognition networks, which will respectively measure the face detection rate and identity similarity of the adversarial images in test. We select pre-trained Resnet-50 and Inception-V3 to calculate the loss of facial attributes and expression respectively. Note that the test dataset is FFHQ at 256x256, which all the models used in the following part are not trained on it. 

\begin{figure}
\setlength{\abovecaptionskip}{0cm} 
\setlength{\belowcaptionskip}{0cm}
\centering
\includegraphics[scale=0.38]{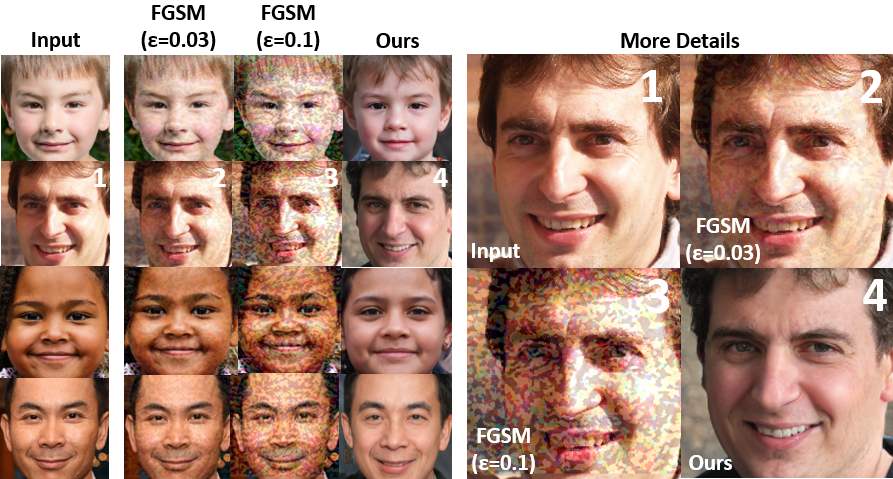}
\caption{The result of qualitative comparisons between FGSM and ours.}
\label{Figure 2}
\end{figure}

\subsection{De-ID by adversarial attacks}
In order to evaluate the adversarial performance of our model, we select the FGSM as the comparison. The step sizes of 0.03 and 0.1 are selected respectively, as shown in the Fig.\ref{Figure 2}. The previous traditional adversarial methods exchange low face detection rate for low identity recognition rate, which have a great impact on the original data distribution. And if you need to keep a low visual artifacts rate, the step size needs to be set very small, but it will degrade its adversarial performance. The image generated by our model has obvious advantages in quality, and can also maintain a high visual ID-similarity and a low model ID-similarity against the original image, which is suitable for De-ID tasks. The quantitative comparative results of the two methods are shown in Table \ref{Table 1}. The visual artifacts rate is measured by MOS experiment: We select 100 images randomly with equal probability 12 times (including original images and De-ID images generated by the three methods), and invite 12 people to judge whether there are any artificial De-ID operations. The influence on other applications of De-ID operation is measured by face detection rate. The cosine similarity with the original ID is used to measure the effect of the method on the level of disturbing ID. The FID metric is used as an objective index to determine the visual ID loss between the original images and the adversarial images.

\begin{table}  
    \centering
    \setlength{\abovecaptionskip}{0cm} 
    \setlength{\belowcaptionskip}{0cm}
    \caption{Quantitative comparison results of FGSM and our method.} 
    \setlength{\tabcolsep}{1mm}
    \begin{tabular}{ccccc} 
    \hline 
    Method  & Artifacts rate $\downarrow$  & Face detect rate $\uparrow$  & ID similarity $\downarrow$  & FID $\downarrow$\\  
    \hline 
    FGSM($\epsilon$=0.03) & 0.9946±0.014 & 0.999 & 0.842±0.005 & \textbf{33.494±0.653}\\
    FGSM($\epsilon$=0.1) & 1 & 0.989 & 0.603±0.008 & 145.923±2.276\\
    Ours & \textbf{0.1442±0.173} & \textbf{1} & \textbf{0.531±0.006} & 61.438±0.533\\
    \hline
    \end{tabular}
    \label{Table 1}
\end{table}  

\subsection{De-ID by face swapping}

\subsubsection{Disentanglement: different-input-and-face-swap}

\begin{figure}
\setlength{\abovecaptionskip}{0cm} 
\setlength{\belowcaptionskip}{0cm}
\centering
\includegraphics[scale=0.31]{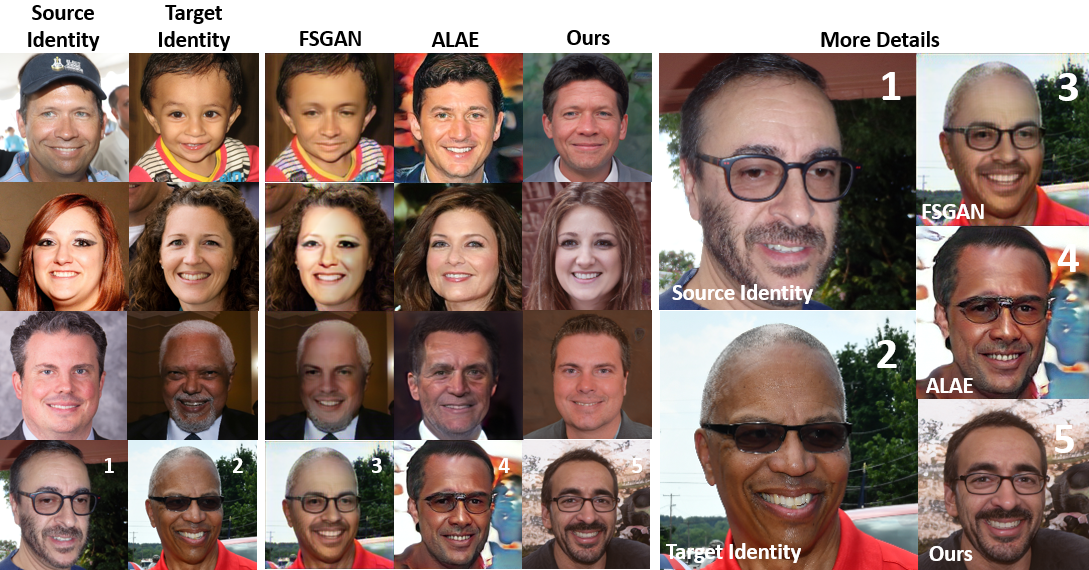}
\caption{The qualitative results of ALAE, FSGAN, and ours on disentanglement.}
\label{Figure 3}
\end{figure}

Compared with ALAE and FSGAN, the qualitative results of the three methods are shown in Fig.\ref{Figure 3}. Both ALAE and our method use the whole image generation method, but our method is superior to ALAE in the preservation of identity, expression and pose. At the same time, our method can control the region outside the face and reduce artifacts, better than ALAE. FSGAN focuses on the face region, so it can preserve the features of target identity well. However, the quality of the image generated by FSGAN is obviously poor, and it cannot make the pose and expression provided by target identity, as effectively as our method. This shows that our method can disentangle between visual identity and expressions well. We select FID as a metric to measure the similarity between the original image and the new image. Finally, we evaluated the retention of identity and expression for the model, and the experimental results are summarized in Table \ref{Tabel 2}.

\begin{table}  
    \setlength{\abovecaptionskip}{0cm} 
    \setlength{\belowcaptionskip}{0cm}
    \centering
    \caption{Evaluation of ALAE, FSGAN and our method on disentanglement.} 
    \setlength{\tabcolsep}{0.5mm}
    \begin{tabular}{cccc} 
    \hline 
    Method & FID $\downarrow$ & Source ID identity loss $\downarrow$ & Target ID expression loss $\downarrow$\\  
    \hline 
    ALAE & 69.632±1.803	 & 0.033 & 2.165±0.018 \\
    FSGAN & 102.324±3.429  & 0.028 & \textbf{0.958±0.038} \\
    Ours & \textbf{37.780±0.100}  & \textbf{0.014} & 1.320±0.012\\
    \hline
    \end{tabular}
    \label{Tabel 2}
\end{table}  

\subsubsection{Reconstruction: same-input-and-face-reconstruct}
\begin{figure}
\setlength{\abovecaptionskip}{0cm} 
\setlength{\belowcaptionskip}{0cm}
\centering
\includegraphics[scale=0.31]{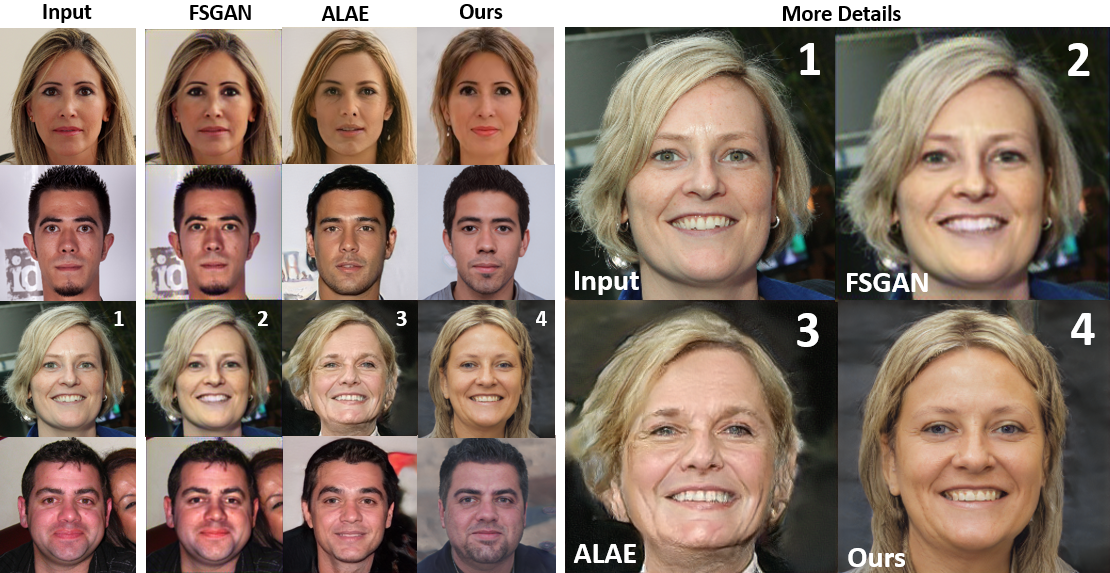}
\caption{The qualitative results of ALAE, FSGAN, and ours on reconstruction.}
\label{Figure 4}
\end{figure}

When $I_{id}$ and $I_{expr}$ are the same, attribute disentanglement and generative network needs to complete the function of reconstruction, so our model is faced with the problem of GAN inversion\cite{xia2021gan}. We compare our method with ALAE and FSGAN, to study the quality of reconstruction of our method. Some visual results are shown in Fig.\ref{Figure 4}. In horizontal contrast, FSGAN, which only focuses on the face, has better performance in the task of reconstruction, but its quality of the generated image is degraded. Both our method and ALAE have undertaken the task of generating the whole image, but our method can better complete GAN inversion, which can preserve most of the features of the original image and reduce the artifacts at the same time. We verified the reconstruction ability of the model from FID, LPIPS\cite{zhang2018unreasonable} and semantic information such as identity and expression, as shown in Table \ref{Table 3}.

\begin{table}  
    \setlength{\abovecaptionskip}{0cm} 
    \setlength{\belowcaptionskip}{0cm}
    \centering
    \caption{Evaluation of ALAE, FSGAN and our method on reconstruction.} 
    \setlength{\tabcolsep}{1mm}
    \begin{tabular}{cccccc} 
    \hline 
    Method & FID $\downarrow$ &LPIPS $\uparrow$ & Identity loss $\downarrow$ & Expression loss $\downarrow$\\  
    \hline 
    ALAE & 77.270±1.852 & \textbf{0.535±0.003}  & 0.032 & 1.491±0.014\\
    FSGAN & 96.010±5.027 & 0.260±0.003  & \textbf{0.008} & \textbf{0.650±0.034}\\
    
    Ours & \textbf{32.889±0.374} & 0.440±0.004 & 0.012 & 0.869±0.008\\
    \hline
    \end{tabular}  
    \label{Table 3}
\end{table} 

\section{Conclusion}
We propose a systematical solution for various levels of de-identification, and the processed images have high image quality. We compare our method with several SOTA methods in specific tasks like adversarial attacks, disentanglement and reconstruction. The experimental results show that our method has achieved good results in these tasks. It is worth mentioning that our method has better generalization than other methods when being tested in non-training dataset. When generating the adversarial sample of the latent vector of the target image, the gradient is back propagated to a mapping network of the adversarial vector, so as to characterize the common features of the adversarial samples. This is to explore a general and unrestricted adversarial sample generation way for face images. However, because the expressive ability of StyleGAN in latent space is limited, we cannot reconstruct the region outside the face. Combined with GAN inversion and unrestricted adversarial attacks, our further research will pay attention to the mechanism of GANs and apply it to De-ID problem of face data.

\section{Acknowledgements}
This work was supported by the National Natural Science Foundation of China 61772529, Beijing Natural Science Foundation under Grant 4192058, National Natural Science Foundation of China 61972395 and National Key Research and Development Program of China 2020AAA0140003.

\begin{sloppypar}
    \printbibliography{}
\end{sloppypar}
\end{document}